# Deletion-Robust Submodular Maximization at Scale

**Ehsan Kazemi**                                    EHSAN.KAZEMI@YALE.EDU
*Yale University, New Haven, CT 06511, USA*

**Morteza Zadimoghaddam**                           ZADIM@GOOGLE.COM
*Google Research, New York, NY 10011, USA*

**Amin Karbasi**                                    AMIN.KARBASI@YALE.EDU
*Yale University, New Haven, CT 06511, USA*

## Abstract

Can we efficiently extract useful information from a large user-generated dataset while protecting the privacy of the users and/or ensuring fairness in representation. We cast this problem as an instance of a deletion-robust submodular maximization where part of the data may be deleted due to privacy concerns or fairness criteria. We propose the first memory-efficient centralized, streaming, and distributed methods with constant-factor approximation guarantees against *any* number of adversarial deletions. We extensively evaluate the performance of our algorithms against prior state-of-the-art on real-world applications, including (i) Uber-pick up locations with location privacy constraints; (ii) feature selection with fairness constraints for income prediction and crime rate prediction; and (iii) robust to deletion summarization of census data, consisting of 2,458,285 feature vectors.

## 1. Introduction

It has long been known that solutions obtained from optimization methods can demonstrate striking sensitivity to the parameters of the problem (Bertsimas et al., 2011). Robust optimization, in contrast, is a paradigm in the mathematical programming community with the aim of safeguarding the solutions from the (bounded) changes in the underlying parameters.

In this paper, we consider submodular maximization, a very well studied discrete optimization problem defined over a finite set of items (e.g., images, videos, blog posts, sensors, etc). Submodularity formalizes the notion of diminishing returns, stating (informally) that selecting an item earlier results in a higher utility than selecting it later. This notion has found far-reaching applications in machine learning (Bach et al., 2013), web search and mining (Borodin et al., 2017), social network (Kempe et al., 2003), crowdsourcing (Singla et al., 2016), and user modeling (Yue and Guestrin, 2011), to name a few. However, almost all the existing methods for submodular maximization, ranging from centralized (Nemhauser et al., 1978; Feldman et al., 2017) to streaming (Badanidiyuru et al., 2014), to distributed (Mirzasoleiman et al., 2013; Mirrokni and Zadimoghaddam, 2015; Barbosa et al., 2015), rely on greedy selection of elements. As a result, the returned solution of such methods are remarkably sensitive to even a *single* deletion from the set of items.

The need for efficient deletion-robust optimization methods is wide-spread across many data-driven applications. With access to big and massive data (usually generated by millions of users), along with strong machine learning techniques, many service providers have been able to exploit these new resources in order to improve the accuracy of their data analytics. At the same time, it has been observed that many such inference tasks may leak very sensitive information about the data providers (i.e., personally identifiable information, protected health information, legal or financial data, etc). Similarly these algorithms can encode hidden biases that disproportionately and adversely impact members with certain characteristics (e.g., gender, race, religion, sexual orientation).

In order to reduce the effect of information extraction on privacy or fairness, one needs to be able to remove sensitive data points (e.g., geolocations) or discard sensitive data features (e.g., skin color) from the dataset without incurring too much loss in performance. For instance, European Commission disclosed a draft "European Data Protection Regulation" that contains concrete guidelines for providing individuals with the "right to be forgotten". By exercising this right, individuals may enforce the service provider to delete their personal data or put restrictions from using part of it. Similarly, Title VII of the Civil Rights Act of American anti-discrimination law prohibits employment discrimination against certain characteristics (such as color and sex). Thus, to obtain fairer machine learning algorithms, we need to reduce the bias inherent in the training examples due to the lack of certain types of information, not being representative, or reflecting historical biases. This can be done by either removing protected attributes from training data (Zemel et al., 2013) or train them separately for different protected groups (Chayes, 2017), among other procedures. Unfortunately, sensitive features or biased data usually are not known a priori and we might get aware of their existence just after training our models (Beutel et al., 2017). Retraining a machine learning model from scratch, after removing sensitive features and biased data, is quite expensive for large datasets. Deletion-robust submodular maximization can save a lot of time and computational resources in these scenarios.

Most existing submodular maximization methods, often used for data extraction (Mirzasoleiman et al., 2013) and informative subset selection (Wei et al., 2015), do not provide such guarantees. In this paper, we develop the first scalable and memory-efficient algorithms for maximizing a submodular function subject to a cardinality constraint that are robust against any number of *adversarial* deletions. This is in sharp contrast to previous methods that could only handle a fixed number of deletions (Orlin et al., 2016; Bogunovic et al., 2017) or otherwise their memory requirement scales multiplicatively with the number of deletions (Mirzasoleiman et al., 2017) (e.g., to handle 10 deletions, the robust solution becomes 10 folds larger).

**Our contributions:** To maximize a monotone submodular function with a cardinality constraint $k$, we develop the following algorithms that are robust against any $d$ adversarial deletions:

- Centralized solution: We propose ROBUST-CENTRALIZED that achieves $(1/2 - \delta)$-approximation guarantee (in expectation) with the memory requirement $O\left(k + (d \log k)/\delta^2\right)$.

- Streaming solution: We propose ROBUST-STREAMING that achieves $(1/2-\delta)$-approximation guarantee (in expectation) with the memory requirement $O\left(\frac{k \log k}{\delta} + d\frac{\log^2 k}{\delta^3}\right)$.

Table 1: Comparison of algorithms for robust monotone submodular maximization with a cardinality constraint.

| Algorithm | Max. Robustness | Approx. | Memory | Setup |
| --- | --- | --- | --- | --- |
| OSU (Orlin et al., 2016) | $o(\sqrt{k})$ | 0.387 | $k$ | Centralized |
| PRO-GREEDY (Bogunovic et al., 2017) | $o(k)$ | 0.387 | $k$ | Centralized |
| ROBUST (Mirzasoleiman et al., 2017) | arbitrary $d$ | $1/2 - \delta$ | $O(kd \log k/\delta)$ | Streaming |
| ROBUST-CENTRALIZED (ours) | arbitrary $d$ | $1/2 - \delta$ | $O(k + d \log k/\delta^2)$ | Centralized |
| ROBUST-STREAMING (ours) | arbitrary $d$ | $1/2 - \delta$ | $O(k \log k/\delta + d \log^2 k/\delta^3)$ | Streaming |
| ROBUST-DISTRIBUTED (ours) | arbitrary $d$ | $0.218 - \delta$ | $O(m(k + d \log k/\delta^2))$ | Distributed |
| COMPACT-DISTRIBUTED (ours) | arbitrary $d$ | $0.109 - \delta$ | $O(k + d \log k/\delta^2)$ | Distributed |

- Distributed solution: We propose ROBUST-DISTRIBUTED that achieves $(0.218 - \delta)$-approximation guarantee (in expectation) with the memory requirement $O\left(m(k + d\frac{\log k}{\delta^2})\right)$, where $m$ is the number of machines. We also introduce COMPACT-DISTRIBUTED, a variant of ROBUST-DISTRIBUTED, where its memory requirement is independent of number of machines.

Table 1 compares our proposed methods with previous algorithms. Our experimental results demonstrate the effectiveness of our proposed methods on several real-life applications.

## 2. Related Work

Monotone submodular maximization under cardinality constraints is studied extensively in centralized, streaming and distributed scenarios. The classical result of Nemhauser et al. (1978) proves that the simple GREEDY algorithm that starts with an empty set and iteratively adds elements with the highest marginal gain provides $(1 - 1/e)$-approximation guarantee. To scale to large datasets, several multi-pass and streaming algorithms with constant factor approximations have recently been proposed (Krause and Gomes, 2010; Badanidiyuru et al., 2014; Kumar et al., 2015; Buchbinder et al., 2015). Also, different distributed submodular maximization algorithms have been developed lately (Mirzasoleiman et al., 2013; Mirrokni and Zadimoghaddam, 2015; Barbosa et al., 2015).

Krause et al. (2008) introduced the robust formulation of the classical cardinality constrained submodular maximization for the first time and gave a bi-criterion approximation to the problem of $\max_{|A| \le k} \min_{i \in \{1, \cdots, \ell\}} f_i(A)$, where $f_i$ is normalized monotone submodular for every $i$. Note that submodular maximization of function $f$ that is robust to the deletion of $d$ items can be modeled as a special case of this problem: $\max_{|A| \le k} \min_{|D| \le d} f(A \setminus D)$. Krause et al. (2008) guarantee a robust solution by returning a set whose size is $k(1+\Theta(\log(dk \log n))$. There are two main drawbacks with this approach when applied to deletions: first, the size of final solution is logarithmically larger than $k$, and second, the running time is exponential in $d$. Orlin et al. (2016) designed a centralized algorithm that outputs a set of cardinality $k$ in a polynomial time. Their algorithm is robust to the deletion of only $o(\sqrt{k})$ elements. Bogunovic et al. (2017) further improved the result of Orlin et al. (2016) to $o(k)$ deletions. The approximation guarantees for both of these algorithms are 0.387. The aforementioned methods try to construct a solution without allowing to update the answer after deletion. In contrast, Mirzasoleiman et al. (2017) developed a streaming algorithm which is robust to the deletion of any number of $d$ elements. They keep a set of size $O(kd \log k/\delta)$, and after each deletion they find a feasible solution of size at most $k$ from this set. They also improved the approximation guarantee to $1/2 - \delta$. The main drawback of this algorithm is the memory requirement, which is quite impractical for large values of $d$ and $k$; e.g., for $k = O(\sqrt{n})$ and $d = O(\sqrt{n})$ the memory requirement is even larger than $n$.

Submodular maximization under a cardinality constraint has been widely used in classical machine learning and data mining applications, including extracting representative elements with exemplar based clustering (Krause and Gomes, 2010), data summarization through active set selection (Herbrich et al., 2003; Seeger, 2004; Krause and Guestrin, 2005), feature selection (Krause and Guestrin, 2005) and document summarization (Lin and Bilmes, 2011). Furthermore, there are many other areas where robust submodular maximization could be quite useful, including influence maximization (He and Kempe, 2016), personalized image summarization (Lucic et al., 2016), sensor placement (Krause et al., 2008) and robust feature selection (Globerson and Roweis, 2006).

## 3. Problem Definition

Assume we have a set function $f : 2^V \to \mathbb{R}_{\geq 0}$. We define the marginal gain of an element $e \in V$ to the set $A \subseteq V$ by $\Delta_f(e|V) = f(A \cup \{e\}) - f(A)$. The function $f$ is submodular if for all $A \subseteq B \subseteq V$ and $e \in V \setminus B$, we have $\Delta_f(e|A) \geq \Delta_f(e|B)$. A submodular function $f$ is monotone if for every $A \subseteq B \subseteq V$, we have $f(A) \leq f(B)$.

Through this paper our focus is on maximizing a monotone submodular function with a cardinality constraint given a parameter $k$, i.e., our goal is to solve the following problem:

$$S^* = \underset{S \subseteq V, |S| \leq k}{\arg\max} f(S). \tag{1}$$

In many applications, a subset of items of the ground set $V$ may be removed and we need to solve Eq. (1) again without the deleted items. For this reason, we require to make our solution robust to the deletion. Indeed, the goal is to maximize a submodular function $f$ over a set $V$ of items, where it is robust to the deletion of any subset $D \subset V$ of size $|D| \leq d$. More precisely, we are interested in solving the following problem for each possible instance of $D$:

$$S^* = \underset{S \subseteq V \setminus D, |S| \leq k}{\arg\max} f(S). \tag{2}$$

We also define $\text{OPT} = f(S^*)$. The most straightforward approach to this problem is to solve Eq. (2) for each instance of $D$. Unfortunately, solving Eq. (2), for large datasets, is computationally prohibitive. Also, deletion of elements from the set $V$ can happen at different stages in real time applications. This makes the problem even harder. Our solution to this problem is to maintain a small subset $A$ of $V$, called a core-set of $V$, where for each set $D$ we can efficiently find a subset $B \subseteq A \setminus D$ that provides an acceptable approximation for Eq. (2). For this reason, next we define the notion of $(\alpha, d)$-robust randomized core-set.

**Definition 1** *A random subset of $A \subseteq V$ is an $(\alpha, d)$-robust randomized core-set for a set $V$, if for any subset $D \subseteq V$ of size $|D| \leq d$, there exists a $B \subseteq A \setminus D, |B| \leq k$ such that*

$$\mathbb{E}[f(B)] \geq \alpha \max_{S \subseteq V \setminus D, |S| \leq k} f(S).$$

## 4. Robustness and Cardinality Constraint

In this section, we present three fast and scalable randomized algorithms. These algorithms solve the problem of robust submodular maximization in centralized, streaming and distributed scenarios. Our algorithms provide, in expectation, constant factor approximation guarantees, where they are robust to the adversarial deletion of any $d$ items from the set $V$. In our setting, an adversary might try to find a set of inputs for which our algorithms fail to provide good results. In order to make the optimization robust to the adversarial deletions, we introduce randomness in the selection process. We also assume that the adversary dose not have access to the random bits of the randomized algorithms.

The algorithms are designed based on a general idea that the elements are chosen randomly from a large enough pool of *similar* items. This idea is useful because the adversary is not aware of the random bits of the algorithms, which makes the deletion probability of elements we have chosen negligible. Therefore, we can bound the expected value of the selected set.

Our solution consists of two steps. In the first step, we find a small core-set of elements (in comparison to the whole dataset). We prove that after the deletion of at most $d$ arbitrary elements, we can still find a good approximation for the optimization problem in this small set. In the second step, we choose at most $k$ elements from the core-set we have found in the first step. We prove a constant approximation factor for our algorithm in expectation. This guarantees that the core-set is $(\alpha, d)$-robust randomized for a constant $\alpha$ and arbitrary $d$.

In the optimization procedure, we use a thresholding idea to select elements. Similar ideas have been used previously for designing streaming algorithms (Badanidiyuru et al., 2014; Buchbinder et al., 2015). In those algorithms, when an element of the stream arrives, if this element has *sufficiently* large marginal value it is kept. Otherwise it is discarded. In the robust submodular maximization, we keep a large enough pool of elements with sufficient marginal values before adding or discarding them. We randomly pick an element when the size of pool is at least $d/\epsilon$. Thus the element picked at each step is deleted with a probability at most $\epsilon$. This is true because the size of deleted items is at most $d$. To guarantee the quality of the chosen elements after the deletion (i.e., we want the expected value of $f$ over the set of picked elements does not change a lot after deletion), not only they should have been picked from a large pool of elements, the elements of pool should have almost the same marginal gains. To explain why we need this property consider the following example.

**Example 1** *Suppose the ground set $V$ consists of identical elements with equal value of 1. In other words, for any $A \subseteq V$, let $f(A)$ be 1 if $A$ is not empty, and let it be 0 for the empty set. In this case, all elements are good candidates to be chosen at the beginning of algorithm. However after choosing any of them, the marginal gain of the rest becomes 0, and the algorithm has no incentive to continue selecting elements. If the first element is chosen deterministically, the adversary can delete that element and we can not find any non-zero value subset after deletion. Now if we pick $d/\epsilon$ of these elements and then pick one of them randomly, the probability that adversary can delete the chosen element reduces to $\epsilon$ and we achieve the robustness we aim for.*

### 4.1 Centralized Algorithm

In this section we outline a centralized algorithm, called ROBUST-CORESET-CENTRALIZED, to find an $(\alpha.d)$-robust core-set for $V$. We also present the ROBUST-CENTRALIZED algorithm which is able to find a good solution for Eq. (2) from the core-set. We use the threshold based algorithmic ideas of Badanidiyuru et al. (2014) that yield the state of the art streaming algorithms for submodular maximization. We then elaborate on the challenges the adversarial deletion process causes for this approach and how we resolve them.

In their work, Badanidiyuru et al. (2014) prove that choosing elements with marginal gain at least $\tau^* = \frac{\text{OPT}}{2k}$ from a stream until a maximum of $k$ elements are chosen returns a set $S$ with an approximation factor of $1/2$. The main problem with this primary idea is that the value of OPT is not known by the algorithm. Badanidiyuru et al. (2014) pointed out that, from the submodularity of $f$, we have $\Delta_0 \leq \text{OPT} \leq k\Delta_0$ where $\Delta_0$ is the largest value in set $\{f(\{e\})|e \in V\}$. By dividing the range $[\Delta_0, k\Delta_0]$ into intervals of $[\tau_i, \tau_{i+1})$ (where $\tau_{i+1}/\tau_i$ is close to 1) it is possible to find a good enough approximation for OPT. Since the ratio of upper and lower limits of the range $[\Delta_0, k\Delta_0]$ is $k$, it suffices to try $\log k$ different estimates to get a close match on one of the guesses. In an streaming algorithm, we do not know the maximum value of singletons a priori. Badanidiyuru et al. (2014) showed that it is enough to look at the range $m_t \leq \text{OPT} \leq 2km_t$, where $m_t$ is the maximum value of elements observed at time $t$.

We first note that due to the deletion process, the relevant maximum singleton value is not $\Delta_0$, it is $\Delta'_0 = \max_{e \in V \setminus D} f(\{e\})$. The algorithm is unaware of set $D$, therefore $\Delta'_0$ could be anywhere in the range $[\Delta_d, \Delta_0]$ where $\Delta_d$ is the $d+1$-th largest value in the set $\{f(\{e\})|e \in V\}$. The lower bound of $\Delta_d$ is implied by the fact that at most $d$ elements will be deleted. So $\tau^* = \frac{\text{OPT}}{2k}$ could fall anywhere in the range $[\Delta_d/2k, \Delta_0]$. Unlike (Badanidiyuru et al., 2014), the upper and lower limits of this range do not differ only by a multiplicative factor of $k$, thus a naive approach makes us try arbitrarily large number of different choices to find a good estimate of $\tau^*$. We resolve this issue by the following observation.

We reserve a set $B$ of elements that might be valuable after the deletion process. Let $V_d$ be the $(d+1)$ largest singleton value elements, i.e., the top $d+1$ elements $e$ in the set $\{f(\{e\})|e \in V\}$. We preserve all elements of $V_d$ for the next round by inserting them to $B$. This way we do not have to worry about thresholds above $\Delta_d$ as all elements that might have marginal value above $\Delta_d$ to any set should be in set $V_d$ and they are added to $B$. Therefore, we consider all thresholds in the set $\text{T} = \{(1+\epsilon)^i | \frac{\Delta_d}{2k} \leq (1+\epsilon)^i \leq \Delta_d\}$.

Starting from the largest $\tau \in \text{T}$ to the smallest threshold in T, we iteratively construct two sets $A_\tau$ and $B_\tau$. At the end of the algorithm, the set $B$ is defined as the union of $V_d$ and $\cup_{\tau \in \text{T}} B_\tau$. We output the set $B$, along with all sets $\{A_\tau\}_{\tau \in \text{T}}$, as the core-set.

We initialize $A_\tau$ to $\emptyset$. We let $B_\tau$ to be the set of elements whose marginal values to the set $\cup_{\tau' \geq \tau} A_{\tau'}$ is in the range $[\tau, (1+\epsilon)\tau)$. We note that this is a dynamic definition and whenever we add an element to any $A_\tau$ set, the related $B_\tau$ set might change as well. Elements in the set $B_\tau$ are similar to each other in terms of their marginal values. Without deletions, we can choose any element from $B_\tau$ and add it to our solution. However, if $B_\tau$ has only a few elements, the adversary can delete all of them, and we will be left with an arbitrary poor solution. To make the selection process robust, we select a random element from $B_\tau$ and add it to $A_\tau$ only if there are at least $d/\epsilon$ elements in $B_\tau$. This way even if all deleted elements are from the set $B_\tau$, the probability of the selected elements being deleted is at most $\epsilon$. We also know that all elements added to $A_\tau$ have similar marginal values and are interchangeable. We keep adding elements to $A_\tau$ until either $\cup_{\tau' \geq \tau} A_{\tau'}$ has $k$ elements or the size of set $B_\tau$ becomes smaller than $d/\epsilon$. At this stage, we keep both sets $A_\tau$ and $B_\tau$ as a part of the output core-set. We also remove them from the ground set $V$ and move on to the next lower threshold. The pseudo code of Robust-Coreset-Centralized is given in Algorithm 1.

---

**Algorithm 1:** Robust-Coreset-Centralized

1 $\Delta_d \leftarrow$ the $(d+1)$-th largest value of set $\{f(\{e\}) | e \in V\}$;
2 $V_d \leftarrow$ all the $d+1$ elements with the largest values in set $\{f(\{e\}) | e \in V\}$;
3 $T = \{(1+\epsilon)^i | \frac{\Delta_d}{2(1+\epsilon)k} \leq (1+\epsilon)^i \leq \Delta_d\}$;
4 For each $\tau \in T : \{A_\tau\} \leftarrow \emptyset$ and $\{B_\tau\} \leftarrow \emptyset$;
5 $V \leftarrow V \setminus V_d$;
6 **for** $\tau \in T$ *from the highest to the lowest* **do**
7     **while** $|B_\tau| \geq \frac{d}{\epsilon}$ for $B_\tau = \{e \in V : \tau \leq \Delta_f(e| \cup_{\tau' \geq \tau} A_{\tau'}) < (1+\epsilon)\tau\}$ *and* $|\cup_{\tau' \geq \tau} A_{\tau'}| < k$ **do**
8         Randomly pick an element $e$ from $B_\tau$ and add it to $A_\tau$, i.e., $A_\tau \leftarrow A_\tau \cup \{e\}$;
9     $V \leftarrow V \setminus (A_\tau \cup B_\tau)$;
10 $B \leftarrow \{\cup B_\tau\} \cup V_d$;
11 **return** $\{A_\tau\}, B$

---

The sets $\{A_\tau\}$ and $B$ are the outputs (core-set) of Robust-Coreset-Centralized. Next we show how Robust-Centralized (with pseudo code given in Algorithm 2) returns a solution for submodular maximization problem after the deletion of set $D$. The subsets of $A_\tau$ and $B$ after the deletion of set $D$ are denoted by $A'_\tau$ and $B'$, respectively, i.e., $A'_\tau = A_\tau \setminus D$ and $B' = B \setminus D$. Robust-Centralized uses the sets $\{A'_\tau\}$ and $B'$ in order to find a good solution to the optimization problem of Eq. (2). Robust-Centralized considers all the possible thresholds in the range $[\Delta'_0/(2k), \Delta'_0]$, where $\Delta'_0$ is the largest value in set $\{f(\{e\}) | e \in V \setminus D\}$. We note that at this point, we can compute the value of $\Delta'_0$ because the set of deleted elements are revealed and we also kept all elements in $V_d$ as part of the core-set. For each threshold $\tau$, we can ensure that the marginal gain of elements in $S_\tau = \cup_{\tau' > \tau} A_{\tau'}$ is at least $\tau$. Therefore, we keep them as part of the solution. Next for any element $e \in B'$ the Robust-Centralized algorithm checks if the marginal gain of $e$ to $S_\tau$ is at least $\tau$. If it is, then $e$ is added to $S_\tau$. We do not need to introduce any extra randomness or selection from a large pool of candidates for additional robustness at this point, since the deletions are done already. The final solution is the set with the maximum value $f(S_\tau)$ among all $S_\tau$.

**Theorem 1** *For any $\delta > 0$, by setting $\epsilon = \frac{2\delta}{3}$,* Robust-Coreset-Centralized *and* Robust-Centralized *satisfy the following properties:*

- Robust-Centralized *outputs a set $S$ such that $|S| \leq k$ and $\mathbb{E}[f(S)] \geq (\frac{1}{2} - \delta)\text{OPT}$.*

**Algorithm 2:** ROBUST-CENTRALIZED

**Input:** $\{A'_\tau\}$ and $B'$ // $A'_\tau$ and $B'$ contain elements of $A_\tau$ and $B$ (outputs of ROBUST-CORESET-CENTRALIZED) after deletion.

**Output:** Set $S$ of cardinality at most $k$

1. $\Delta'_0 \leftarrow$ the largest value of set $\{f(\{e\}) | e \in \{\cup A'_\tau\} \cup B'\}$;
2. $T' = \{(1+\epsilon)^i | \frac{\Delta'_0}{2(1+\epsilon)k} \leq (1+\epsilon)^i \leq \Delta'_0\}$;
3. **for** $\tau \in T'$ from the highest to the lowest **do**
4. $\quad S_\tau \leftarrow \bigcup_{\tau' \in T', \tau' \geq \tau} A'_{\tau'}$;
5. $\quad$ **for** all $e \in B'$ **do**
6. $\quad\quad$ **if** $\Delta_f(e|S_\tau) \geq \tau$ and $|S_\tau| < k$ **then**
7. $\quad\quad\quad S_\tau \leftarrow S_\tau \cup e$;
8. $\quad$ **return** $\arg\max_{S_\tau} f(S_\tau)$

- ROBUST-CORESET-CENTRALIZED outputs at most $O\left(k + d\frac{\log k}{\delta^2}\right)$ elements as the core-set. Since this core-set is the input of ROBUST-CENTRALIZED, its space complexity is upper bounded by the same number of elements.

- The query complexities of ROBUST-CORESET-CENTRALIZED and ROBUST-CENTRALIZED are, respectively, $O\left((k + \frac{\log k}{\delta})|V|\right)$ and $O\left((k + d\frac{\log k}{\delta^2})\frac{\log k}{\delta}\right)$.

**Proof** We define $V' = V \setminus D$. Assume $A'_\tau$ and $B'$, respectively, are subsets of $A_\tau$ and $B$ after deletion of set $D$ from $V$. We define $S^* = \arg\max_{S \subseteq V \setminus D, |S| \leq k} f(S)$ and $f(S^*) = \text{OPT}$. We start by showing that one of the thresholds the ROBUST-CENTRALIZED algorithm tries is close to the standard threshold $\frac{\text{OPT}}{2k}$ that guarantees the $\frac{1}{2}$ approximation without deletion.

**Lemma 1** *There is a $\tau^* \in T'$ such that $\tau^* \leq \frac{\text{OPT}}{2k} < \tau^*(1+\epsilon)$, where $T'$ is defined in line 2 of* ROBUST-CENTRALIZED.

**Proof** From the submodularity of $f$ we have $\Delta'_0 \leq \text{OPT} \leq k\Delta'_0$. Therefore, the smallest threshold in $T'$ is at most $\frac{\text{OPT}}{2k}$. Setting $\tau^*$ to be the largest threshold in $T'$ that does not exceed $\frac{\text{OPT}}{2k}$ will satisfy the claim of this lemma. ∎

Since ROBUST-CENTRALIZED tries different thresholds and outputs the maximum value solution among them, it suffices to lower bound the expected value of $f(S_{\tau^*})$ by $(\frac{1}{2} - \delta)\text{OPT}$. We note that $S_{\tau^*}$ consists of two parts: the elements added in the first stage (ROBUST-CORESET-CENTRALIZED) that are not deleted, i.e. $\cup_{\tau \geq \tau^*} A'_\tau$, and the set of elements added in the second stage (line 7 of ROBUST-CENTRALIZED). We start by showing that the effect of deletion on the value of the first part is negligible due to the robustness of how we insert elements in ROBUST-CORESET-CENTRALIZED. To simplify the analysis, we abuse the notation, and define $A = \cup_{\tau \geq \tau^*} A_\tau$ and $A' = \cup_{\tau \geq \tau^*} A'_\tau$.

**Lemma 2** $\mathbb{E}[f(A')] \geq (1-2\epsilon)\mathbb{E}[f(A)]$, *and consequently we have* $\mathbb{E}[f(S_{\tau^*})] \geq (1-2\epsilon)\mathbb{E}[f(S_{\tau^*} \cup A)]$ *where the expectations are taken over the random coin flips of* ROBUST-CORESET-CENTRALIZED.

**Proof** We represent elements of $A_\tau$ with $A_\tau = \{e_{\tau,1}, \cdots, e_{\tau,n_\tau}\}$. Similarly, we define $A'_\tau = \{e'_{\tau,1}, \cdots, e'_{\tau,n'_\tau}\}$. We also define $n_\tau = |A_\tau|$ and $n'_\tau = |A'_\tau|$. We have

$$f(A) = \sum_{\tau=\tau_{max}}^{\tau^*} \sum_{l=1}^{|A_\tau|} \Delta_f(e_{\tau,l} | \cup_{\tau'>\tau} A_{\tau'} \cup \{e_{\tau,1}, \cdots, e_{\tau,l-1}\}),$$

where $\tau_{max}$ is the highest threshold in T'. The marginal gain for all elements of $A_\tau$ is sandwiched in the narrow range $[\tau, (1+\epsilon)\tau]$. Therefore, we can bound the value of $A$ in terms of the sizes of $A_\tau$ sets and their associated thresholds:

$$\sum_{\tau \geq \tau^*} |A_\tau|\tau \leq f(A) \leq (1+\epsilon) \sum_{\tau \geq \tau^*} |A_\tau|\tau.$$

By taking the expected value of each side of these bounds, we get:

$$\sum_{\tau \geq \tau^*} \mathbb{E}[|A_\tau|]\tau \leq \mathbb{E}[f(A)] \leq (1+\epsilon) \sum_{\tau \geq \tau^*} \mathbb{E}[|A_\tau|]\tau. \qquad (3)$$

Each element of $A_\tau$ is picked randomly from a set of size $\frac{d}{\epsilon}$. This means that each of these elements are deleted with a probability at most $\epsilon$. From the submodularity of $f$, we know that the marginal gain of elements of $A'_\tau$ will not decrease after deletion of any other element. Note that we have $A'_\tau \subseteq A_\tau$. Therefore, we can lower bound the expected value of remaining elements, i.e., $f(A')$, similarly:

$$\mathbb{E}[f(A')] = \sum_{\tau=\tau_{max}}^{\tau^*} \sum_{l=1}^{|A_\tau|} \mathbb{E}[\mathbb{I}_{e_{\tau,l} \notin D} \Delta_f(e_{\tau,l} | \cup_{\tau'>\tau} A'_{\tau'} \cup \{\mathbb{I}_{e_{\tau,1} \notin D} e_{\tau,1}, \cdots, \mathbb{I}_{e_{\tau,l-1} \notin D} e_{\tau,l-1}\})]$$

$$\stackrel{(a)}{\geq} \sum_{\tau=\tau_{max}}^{\tau^*} \sum_{l=1}^{|A_\tau|} \mathbb{E}[\mathbb{I}_{e_{\tau,l} \notin D} \Delta_f(e_{\tau,l} | \cup_{\tau'>\tau} A_{\tau'} \cup \{e_{\tau,1}, \cdots, e_{\tau,l-1}\})]$$

$$\geq \sum_{\tau=\tau_{max}}^{\tau^*} \sum_{l=1}^{|A_\tau|} \Pr[e_{\tau,l} \notin D]\tau \geq (1-\epsilon) \sum_{\tau \geq \tau^*} \mathbb{E}[|A_\tau|]\tau, \qquad (4)$$

where $\mathbb{I}_{e \notin D}$ is a binary indicator variable to check $e \notin D$. Inequality $(a)$ is concluded from the submodularity of $f$. By combining Eqs. (3) and (4), we conclude that:

$$\mathbb{E}[f(A')] \geq \frac{1-\epsilon}{1+\epsilon} \mathbb{E}[f(A)] \geq (1-2\epsilon)\mathbb{E}[f(A)].$$

So far we have proved that the expected value of $A'$ is not much smaller than the value of $A$. We note that by definition $A'$ is a subset of both $S_{\tau^*}$ and $A$. By submodularity, we have:

$$f(S_{\tau^*} \cup A) - f(S_{\tau^*}) \leq f(A) - f(A').$$

We have shown that the expected value of the right hand side is at most $2\epsilon\mathbb{E}[f(A)]$ which completes the proof, since $f(A) \leq f(S_{\tau^*} \cup A)$ by monotonicity of $f$. ∎

We have shown that values of $S_{\tau^*}$ and $S_{\tau^*} \cup A$ do not differ by much. So we can focus on lower bounding $f(S_{\tau^*} \cup A)$ in the rest of the proof.

**Lemma 3** $f(S_{\tau^*} \cup A) \geq \frac{(1-\epsilon)\text{OPT}}{2}$.

**Proof** The while loop condition in line 7 of Robust-Coreset-Centralized ensures that there will be at most $k$ elements in $A$. If $A$ has exactly $k$ elements, its value is at least $k\tau^* \geq \frac{\text{OPT}}{2(1+\epsilon)} \geq \frac{(1-\epsilon)\text{OPT}}{2}$, since each element added to $A$ increases its value by some threshold $\tau \geq \tau^*$. Monotonicity of $f$ implies that $f(S_{\tau^*} \cup A) \geq f(A)$ which completes the proof in this case. Similarly, the claim is proved if $S_{\tau^*}$ has $k$ elements. So in the rest of the proof, we focus on the case $|A| < k$ and $|S_{\tau^*}| < k$.

We define $S_{\tau^*,e}$ to be the subset of $S_{\tau^*}$ which is selected by Robust-Centralized exactly before processing $e$. We have

$$f(S^*) \overset{(a)}{\leq} f(S^* \cup S_{\tau^*} \cup A) \overset{(b)}{\leq} f(S_{\tau^*} \cup A) + \sum_{e \in S^* \setminus (S_{\tau^*} \cup A)} f(e | S_{\tau^*} \cup A)$$

$$\overset{(c)}{\leq} f(S_{\tau^*} \cup A) + \sum_{e \in (S^* \setminus (S_{\tau^*} \cup A)) \setminus B'} f(e | A) + \sum_{e \in (S^* \setminus (S_{\tau^*} \cup A)) \cap B'} f(e | S_{\tau^*,e})$$

$$\overset{(d)}{\leq} f(S_{\tau^*} \cup A) + k\tau^* \implies \frac{\text{OPT}}{2} \overset{(e)}{\leq} f(S_{\tau^*} \cup A).$$

Inequality $(a)$ is true because $f$ is monotone. From the submodularity of $f$ we conclude $(b)$. We have $A \subseteq S_{\tau^*} \cup A$ and $S_{\tau^*,e} \subseteq S_{\tau^*} \cup A$. Thus $(c)$ results from the submodularity of $f$.

To prove inequality $(d)$, we first note that the elements $e \in (S^* \setminus (S_{\tau^*} \cup A)) \setminus B'$ are discarded by Robust-Coreset-Centralized. Since $A$ has strictly less than $k$ elements, they were not discarded because of the cardinality constraint. So, for all of them we have $\Delta_f(e|A) < \tau^*$ (low marginal value). Elements $e \in (S^* \setminus (S_{\tau^*} \cup A))$ are not selected by Robust-Centralized, and cardinality constraint was not the reason for their rejection. Therefore, for these elements we have $f(e|S_{\tau^*,e}) < \tau^*$. ∎

From the results of Lemmas 2 and 3, we know $\mathbb{E}[f(S_{\tau^*})]$ is at least $\frac{(1-3\epsilon)\text{OPT}}{2}$ which proves the first claim of this theorem.

The number of thresholds in Robust-Coreset-Centralized is $O(\log k/\epsilon)$. For each threshold $\tau$, we store at most $d/\epsilon$ items in a $B_\tau$ set. Also, the maximum number of elements in $\{\cup A_\tau\}$ is $k$. In addition, we have $d+1$ items in $V_d$. Therefore, the size of core-set returned by Robust-Coreset-Centralized is at most $O\left(k + (d\log k)/\epsilon^2\right)$ elements. For the query complexity of Robust-Coreset-Centralized we have: (i) each element is considered for at most $O(\log k/\epsilon)$ different thresholds, resulting in $O((|V|\log k)/\epsilon)$ oracle evaluations, and (ii) when an element is picked from $B_\tau$ to be added to $A_\tau$, we should re-calculate marginal gain of elements and update $B_\tau$ resulting in $k|V|$ oracle evaluations since the size of the union set $\cup_{\tau \in T} A_\tau$ never exceeds $k$. Robust-Centralized receives the core-set as the input so it only processes $O\left(k + (d\log k)/\epsilon^2\right)$ elements. Each of them is considered to be added to one of the $O(\log k/\epsilon)$ sets $\{S_\tau\}_{\tau \in T'}$ which results in $O\left((k + d\frac{\log k}{\epsilon^2})\frac{\log k}{\epsilon}\right)$ oracle evaluations. ∎

## 4.2 Streaming Algorithm

In many applications, the dataset does not fit in the main memory of a single machine or even the data itself arrives as a stream. So it is not possible to use centralized algorithms which need random access to the whole data. In this section, we present a streaming algorithm with a limited available memory. We first use the thresholding idea of Section 4.1 in order to find a core-set for $V$. Then we show that it is possible to find a good solution from this core-set when deletion happens. Recall that for ROBUST-CORESET-CENTRALIZED, the maximum singleton element and the thresholds are fixed while in the streaming setting, they may change as new elements arrive. To apply ideas of the centralized algorithm, we should overcome the following challenges: (i) it is not possible to make several passes over the data for different thresholds (i.e., we cannot start from the largest possible marginal gain to the lowest), and (ii) the value of $\Delta_0$ and $\Delta_d$ are not known a priori.

We show (similar to (Badanidiyuru et al., 2014)) that it is possible to maintain a good approximation of OPT even with a single pass over the data. From now on, let $\Delta_0$ and $\Delta_d$, respectively, denote the largest and $d+1$-th largest singleton values in the stream of data at time step $t$. First, note that $\Delta_d \leq$ OPT and the marginal gain of all the currently received elements is at most $\Delta_0$. Therefore, it is enough to consider thresholds in the range $[\frac{\Delta_d}{2k}, \Delta_0]$. A new threshold is instantiated when the maximum singleton element is changed. These increasing new thresholds are between the current maximum and the previous one. Therefore, all the elements with marginal gains larger than the new threshold will appear after its instantiation.

ROBUST-CORESET-STREAMING, for each threshold $\tau$, keeps two sets $A_\tau$ and $B_\tau = \cup_{\tau' \geq \tau} B_{\tau,\tau'}$. All the elements with marginal gains at least $\tau$ (marginal gain of an element $e$ is calculated dynamically by adding $e$ to $A_\tau$, i.e, it is $\Delta_f(e|A_\tau)$) are *good enough* to be picked by this instance of the algorithm. In order to make the selected elements robust to deletions, we should put all good enough elements in different $B_{\tau,\tau'}$ sets, with thresholds $\tau'$ in the range $[\tau, \Delta_0]$, based on their marginal values. Whenever a set $B_{\tau,\tau'}$ becomes large, we pick one element of it randomly to add to $A_\tau$. This ensures that an element is picked from a large pool of almost similar elements. Formally, all the elements with a marginal gain in the range $[\tau', \tau'(1+\epsilon))$ are added to the set $B_{\tau,\tau'}$. When the size of a $B_{\tau,\tau'}$ is at least $d/\epsilon$, we randomly pick an element from $B_{\tau,\tau'}$ and add it to $A_\tau$. Adding an element to $A_\tau$ may decrease the marginal gains of elements in $B_{\tau,\tau'}$ sets. So we recompute their marginal gains and put them in the right $B_{\tau,\tau''}$ set (they are kept if their marginal gains are at least $\tau$, otherwise they are discarded). These changes may make another set large, so we keep adding elements to $A_\tau$ while we find a large $B_{\tau,\tau''}$ set. This process continues until a maximum of $k$ elements are added to $A_\tau$ or the stream of data ends. Note that there are at most $d$ elements with marginal gains in the range $(\Delta_d, \Delta_0]$; we can simply keep these elements (refer to it as set $V_d$). For all $\Delta_d < \tau \leq \Delta_0$, we have $A_\tau = \emptyset$, because there is no pool of size at least $d/\epsilon$ elements to pick from it. Also, for $B_{\tau,\tau'}$ sets, we do not need to cover the range $(\Delta_d, \Delta_0]$ with too many thresholds. Indeed, when $\Delta_d$ changes (it can only increase), we can update the set $V_d$ and locate the removed elements from $V_d$ into a correct $B_{\tau,\tau'}$. Therefore, it is sufficient to consider only thresholds in the range $[\frac{\Delta_d}{2k}, \Delta_d]$. The pseudo code of ROBUST-CORESET-STREAMING is given in Algorithm 3.

**Algorithm 3:** ROBUST-CORESET-STREAMING

**1** $T = \{(1+e)^i | i \in \mathbb{Z}\}$;
**2** For each $\tau, \tau' \in T : \{A_\tau\} \leftarrow \emptyset$ and $\{B_{\tau,\tau'}\} \leftarrow \emptyset$;
**3** **for** *every arriving element $e_t$* **do**
**4**     $\Delta_d \leftarrow$ the $(d+1)$-th largest element of $\{f\{e_1\}, \cdots, \{f\{e_t\}\}$ // If $t \leq d+1$, then $\Delta_d$ is the smallest singleton value.
**5**     $\Delta_0 \leftarrow$ the largest element of $\{f\{e_1\}, \cdots, \{f\{e_t\}\}$;
**6**     $T_t = \{(1+\epsilon)^i | \frac{\Delta_d}{2(1+\epsilon)k} \leq (1+\epsilon)^i \leq \Delta_d\}$;
**7**     Delete all $A_\tau$ and $B_{\tau,\tau'}$ such the $\tau$ or $\tau' \notin T_t$;
**8**     **for** $\tau \in T_t$ **do**
**9**       **if** $|A_\tau| < k$ and $\tau \leq \Delta_f(e | A_\tau)$ **then**
**10**         Add $e_t$ to $B_{\tau,\tau'}$ such that for $\tau' \leq \Delta_f(e_t | A_\tau) < \tau'(1+\epsilon)$;
**11**         **while** $\exists \tau''$ *such that* $|B_{\tau,\tau''}| \geq d/\epsilon$ **do**
**12**           Randomly pick an element $e$ from $B_{\tau,\tau''}$ and add it to $A_\tau$, i.e., $A_\tau \leftarrow A_\tau \cup \{e\}$;
**13**           For all $e \in \bigcup_{\tau'' \in T_i, \tau'' \geq \tau} B_{\tau,\tau''}$ recompute $\Delta_f(e | A_\tau)$ and re-place them in correct bins;
**14** **for** $\tau \in T_n$ **do**
**15**     $B_\tau \leftarrow \bigcup_{\tau' \in T_n, \tau' \geq \tau} B_{\tau,\tau'}$;
**16** **return** $\{A_\tau\}, \{B_\tau\}$

At the end of ROBUST-CORESET-STREAMING, we know there is one running instance of the algorithm with a threshold $\tau^*$ such that $\tau^* \leq \frac{\text{OPT}}{2k} < (1+\epsilon)\tau^*$. For all $e \in V \setminus (A_{\tau^*} \cup B_{\tau^*})$, we have $\Delta_f(e | A_{\tau^*}) < \tau^*$. This ensures that the marginal gain of elements that are not picked by this running instance are smaller than $\frac{\text{OPT}}{2k}$. Let $\{A'_\tau\}$ and $\{B'_\tau\}$ be the subsets of $\{A_\tau\}$ and $\{B_\tau\}$ after the deletion of the set $D$ from $V$, respectively. The elements of $A_{\tau^*}$ are robust to the deletion, i.e., $\mathbb{E}[f(A'_{\tau^*})] \geq (1-2\epsilon)\mathbb{E}[f(A_{\tau^*})]$. Also, all the elements with marginal gain of at least $\tau^*$ are kept in the set $B'_{\tau^*}$. Finally, ROBUST-STREAMING, by adding elements of $B'_{\tau^*}$ with a marginal gain at least $\tau^*$ to $A'_{\tau^*}$, finds a solution with an expected approximation guarantee of $\frac{1-3\epsilon}{2}$ to the optimum solution. The pseudo code of ROBUST-STREAMING is given in Algorithm 4.

**Theorem 2** *For any $\delta > 0$, by setting $\epsilon = \frac{2\delta}{3}$, ROBUST-CORESET-STREAMING and ROBUST-STREAMING satisfy the following properties:*

- ROBUST-STREAMING *outputs a set $S$ such that $|S| \leq k$ and $\mathbb{E}[f(S)] \geq (\frac{1}{2} - \delta)\text{OPT}$.*

- ROBUST-CORESET-STREAMING *makes one passes over the data.*

- ROBUST-CORESET-STREAMING *outputs at most $O\left(\frac{k \log k}{\delta} + \frac{d \log^2 k}{\delta^3}\right)$ elements as the core-set. Since this core-set is the input of* ROBUST-STREAMING, *its space complexity is upper bounded by the same number of elements.*

- *The query complexities of* ROBUST-CORESET-STREAMING *and* ROBUST-STREAMING *are, respectively, $O\left(\frac{\log k}{\delta}|V| + \frac{dk \log^2 k}{\delta^3}\right)$ and $O\left(\frac{d \log^3 k}{\delta^4}\right)$.*

**Algorithm 4:** ROBUST-STREAMING

**Input:** $\{A'_\tau\}$ and $\{B'_\tau\}$ // $A'_\tau$ and $B'_\tau$ contain elements of $A_\tau$ and $B_\tau$ (outputs of ROBUST-CORESET-STREAMING) after deletion.
**Output:** Set $S$ of cardinality at most $k$

1 $\Delta'_0 \leftarrow$ the largest value of set $\{f(\{e\})|e \in \{\cup A'_\tau\} \cup \{\cup B'_\tau\}\}$;
2 $\mathrm{T}' = \{(1+\epsilon)^i | \frac{\Delta'_0}{2(1+\epsilon)k} \leq (1+\epsilon)^i \leq \Delta'_0\}$;
3 **for** $\tau \in \mathrm{T}'$ **do**
4 $\quad$ $S_\tau \leftarrow A'_\tau$;
5 $\quad$ **for** all $e \in B'_\tau$ **do**
6 $\quad\quad$ **if** $|S_\tau| < k$ and $\Delta_f(e|S_\tau) \geq \tau$ **then**
7 $\quad\quad\quad$ $S_\tau \leftarrow S_\tau \cup e$;
8 **return** $\arg\max_\tau f(S_\tau)$

**Proof** The proof is similar to the proof of Theorem 1. We define $V' = V \setminus D$. Assume $A'_\tau$ and $B'_\tau$, respectively, are subsets of $A_\tau$ and $B_\tau$ after deletion of set $D$ from $V$. We define

$$S^* = \underset{S \subseteq V \setminus D, |S| \leq k}{\arg\max} f(S) \text{ and } f(S^*) = \mathrm{OPT}.$$

In our proof, we should consider three points. First, there is a $\tau^* \in \mathrm{T}'$ such that $\tau^* \leq \frac{\mathrm{OPT}}{2k} < \tau^*(1+\epsilon)$. Second, we can show that $\mathbb{E}[f(A'_{\tau^*})] \geq (1-2\epsilon)\mathbb{E}[f(A_{\tau^*})]$. Third, all the elements with enough marginal gain are in the set $B'_{\tau^*}$ and ROBUST-CENTRALIZED will add them to the final solution.

**First** Note that $\Delta'_0 \leq \mathrm{OPT} \leq k\Delta'_0$ and $\mathrm{T}'$ contains all the thresholds in $[\frac{\Delta'_0}{2(1+\epsilon)k}, \Delta'_0]$. Also, $\Delta_d \leq \Delta'_0 \leq \Delta_0$. Therefore, there is a threshold $\tau^*$ such that $\tau^* \leq \frac{\mathrm{OPT}}{2k} < \tau^*(1+\epsilon)$ and it is in both $\mathrm{T}'$ and $\mathrm{T}_n$.

**Second** For the threshold $\tau^*$, ROBUST-CORESET-STREAMING returns two sets $A_{\tau^*}$ and $B_{\tau^*}$, where $B_{\tau^*}$ is the union of sets $B_{\tau^*,\tau}$. Assume $A_{\tau^*}$ has $n_{\tau^*}$ elements and out of these $n_{\tau^*}$ elements, $n_{\tau^*,\tau}$ elements are picked from $B_{\tau^*,\tau}$. This means their marginal gain is in the range of $[\tau, \tau(1+\epsilon)]$. We can bound $f(A_{\tau^*})$ from above by

$$\sum_{\tau \geq \tau^*} n_{\tau^*\tau}\tau \leq f(A_{\tau^*}) \leq (1+\epsilon)\sum_{\tau \geq \tau^*} n_{\tau^*\tau}\tau$$

By taking the expected value of each side of these bounds, we get:

$$\sum_{\tau \geq \tau^*} \mathbb{E}[n_{\tau^*\tau}]\tau \leq \mathbb{E}[f(A_{\tau^*})] \leq (1+\epsilon)\sum_{\tau \geq \tau^*} \mathbb{E}[n_{\tau^*\tau}]\tau \quad (5)$$

We know that an element which is picked at a given step is deleted with a probability at most $\epsilon$. The expected number of elements picked from $B_{\tau^*,\tau}$ that remains in the set $A'_\tau$ (set $A_\tau$ after deletion) is $\mathbb{E}[n'_{\tau^*,\tau}] \geq (1-\epsilon)\mathbb{E}[n_{\tau^*,\tau}]$. Due to the submodularity of $f$, the marginal gain of these undeleted elements is at least $\tau$. To sum up, we have

$$\mathbb{E}[f(A'_\tau)] \geq (1-\epsilon)\sum_{\tau \geq \tau^*} \mathbb{E}[n_{\tau^*,\tau}]\tau.$$

Therefore, we have

$$\mathbb{E}[f(A'_{\tau^*})] \geq \frac{1-\epsilon}{1+\epsilon}\mathbb{E}[f(A_{\tau^*})] \geq (1-2\epsilon)\mathbb{E}[f(A_{\tau^*})] \tag{6}$$

Let $S_{\tau^*}$ denote the set returned by Robust-Streaming for threshold $\tau^*$. To prove $\mathbb{E}[f(S_{\tau^*})] \geq (\frac{1}{2} - 3\epsilon)\text{OPT}$, we consider three cases. If $|A_{\tau^*}| = k$, then $\mathbb{E}[f(A_{\tau^*})] \geq k\tau^* \geq \frac{\text{OPT}}{2(1+\epsilon)} \geq (1-\epsilon)\frac{\text{OPT}}{2}$ and from Eq. (6) we have $\mathbb{E}[f(S_{\tau^*})] \geq \mathbb{E}[f(A'_{\tau^*})] \geq (1-2\epsilon)\mathbb{E}[f(A_{\tau^*})] \geq \frac{(1-3\epsilon)\text{OPT}}{2}$. The claim is proved similarly if $S_{\tau^*}$ has $k$ elements. Let's assume $|A_{\tau^*}| < k$ and $|S_{\tau^*}| < k$.

**Lemma 4** $\mathbb{E}[f(S_{\tau^*})] \geq (1-2\epsilon)\mathbb{E}[f(S_{\tau^*} \cup A_{\tau^*})]$. Also if $|A_{\tau^*}| < k$ and $|S_{\tau^*}| < k$, then $f(S_{\tau^*} \cup A_{\tau^*}) \geq \frac{(1-\epsilon)\text{OPT}}{2}$.

The proof of this lemma is similar to the proofs of Lemmas 2 and 3 and we skip the details. To sum-up, for the case $|A_{\tau^*}| < k$, from Lemma 4, we have $\frac{(1-3\epsilon)\text{OPT}}{2} \leq \mathbb{E}[f(S_{\tau^*})]$. This concludes the first claim of theorem.

Number of thresholds in Robust-Coreset-Streaming in the interval $[\frac{\Delta_d}{2(1+\epsilon)k}, \Delta_d]$ is $O(\frac{\log k}{\epsilon})$. For each $\tau$ in this interval, there are $O(\frac{\log k}{\epsilon})$ sets of $B_{\tau,\tau'}$. We store at most $\frac{d}{\epsilon}$ elements in each of $B_{\tau,\tau'}$ set. Also, the maximum number of elements in $A_\tau$ is $k$. Also, there at most $d$ elements with the marginal gain in range $(\Delta_d, \Delta_0)$. To sum up, Robust-Coreset-Streaming stores $O(\frac{\log k}{\epsilon}(k + \frac{d \log k}{\epsilon^2}) + d) = O(\frac{k \log k}{\epsilon} + \frac{d \log^2 k}{\epsilon^3}))$ elements. For the time complexity of Robust-Coreset-Streaming we have: (i) each element is considered in at most $O(\frac{\log k}{\epsilon})$ different thresholds resulting in $O(\frac{\log k}{\epsilon}|V|)$ oracle evaluations, and (ii) for each threshold, when an element is picked from $B_{\tau,\tau'}$ to be added to $A_\tau$, we should re-calculate marginal gains of all elements in $\cup_{\tau'' \geq \tau} B_{\tau,\tau''}$ resulting in $O(\frac{dk \log k}{\epsilon^2})$ oracle evaluations. This is true because, for each $\tau$, the size of $A_\tau$ never exceeds $k$ and we have at most $O(\frac{d \log k}{\epsilon^2})$ elements in $\cup_{\tau'' \geq \tau} B_{\tau,\tau''}$. Therefore, the total time complexity of Robust-Coreset-Centralized is $O(\frac{\log k}{\epsilon}|V| + \frac{dk \log^2 k}{\epsilon^3})$. Robust-Streaming receives the core-set as the input so it only processes $O(\frac{k \log k}{\epsilon} + \frac{d \log^2 k}{\epsilon^3})$ elements. From the input, only $O(\frac{d \log^2 k}{\epsilon^3})$ elements are in $B'_\tau$. Each of them is considered to be added to one of the $O(\frac{\log k}{\epsilon})$ sets $\{S_\tau\}_{\tau \in T'}$ which results in $O(\frac{d \log^3 k}{\epsilon^4})$ oracle evaluations. ∎

### 4.3 Distributed Algorithm

The exponential growth of data makes it difficult to process or even store the data on a single machine. For this reason, there is an urgent need to develop distributed or parallel computing methods to process massive datasets. Classical centralized submodular maximization algorithms are sequential in nature; they require access to the whole dataset and are not amenable to the distributed settings.

Fortunately, there is a simple solution to the problem of distributed submodular maximization within a MapReduce framework (Barbosa et al., 2015). This distributed algorithm first partitions the data randomly onto $m$ machines. Each machine runs the classical GREEDY algorithm over its chunk of data and finds a solution. In the second step, the outputs of GREEDY on all machines are collected in a single machine and another instance of GREEDY is applied to find a new solution. The final solution is the best one among all the collected answers. Mirrokni and Zadimoghaddam (2015), and Barbosa et al. (2015) proved that this algorithm provides a solution with a constant factor approximation guarantee. In this section, we use similar ideas to present a robust distributed submodular maximization algorithm, called ROBUST-DISTRIBUTED. We prove that our distributed algorithm finds an $(\alpha, d)$-robust randomized core-set with a constant $\alpha$ and any arbitrary $d$.

ROBUST-DISTRIBUTED is a two-round distributed algorithm. It first randomly partitions dataset between $m$ machines. Each machine $i$ runs ROBUST-CORESET-CENTRALIZED on its data and passes the result (i.e., sets $\{A_\tau^i\}$ and $B^i$) to a central machine. After the deletion of the set $D$, this single machine runs $m$ instances of ROBUST-CENTRALIZED on the outputs received from each machine $i$ and finds solutions $S^i$. In addition, it runs the classical GREEDY[1] on the union of sets received from all machines (i.e., union of all sets $\{A_\tau^{i'}\}$ and $B^{i'}$) to find another solution $T$. The final solution is the best answer among $T$ and sets $S^i$. ROBUST-DISTRIBUTED is outlined in Algorithm 5.

---
**Algorithm 5:** ROBUST-DISTRIBUTED
---
**1 for** $e \in V$ **do**
**2** $\quad$ Assign $e$ to a machine $i$ chosen uniformly at random;
**3** Let $V_i$ be the elements assigned to machine $i$;
**4** Run ROBUST-CORESET-CENTRALIZED (Algorithm 1) on each machine to obtain $\{A_\tau^i\}$ and $B^i$;
**5** Run ROBUST-CENTRALIZED (Algorithm 2) on each $\{A_\tau^{i'}\}$ and $B^{i'}$ to get the set $S^i$ of cardinality at most $k$ from each machine// $\{A_\tau^{i'}\}$ and $B^{i'}$ are elements of $\{A_\tau^i\}$ and $B^i$ after deletion of set $D$.
**6** $S \leftarrow \arg\max_{S^i}\{f(S^i)\}$;
**7** $T \leftarrow \text{GREEDY}(\{\bigcup_i \bigcup_{\tau \in T^i} A_\tau^{i'}\} \bigcup \{\bigcup_i B^{i'}\})$// Run the GREEDY algorithm over the union of elements returned by ROBUST-CORESET-CENTRALIZED from all machines. $T^i$ is the set of thresholds in machine $i$.
**8 return** $\arg\max\{f(T), f(S)\}$
---

**Theorem 3** *For any $\delta > 0$, by setting $\epsilon = \delta/2$, ROBUST-DISTRIBUTED satisfies the following properties:*

- *It outputs a set $S$ such that $|S| \leq k$ and $\mathbb{E}[f(S)] \geq \frac{\alpha\beta}{\alpha+\beta}\text{OPT}$, where $\alpha = \frac{1}{3} - \delta$ and $\beta = 1 - \frac{1}{e}$. This results in an approximation factor of $0.218 - \delta$.*

- *ROBUST-DISTRIBUTED outputs at most $O\left(m(k + d\frac{\log k}{\delta^2})\right)$ elements as the core-set, where $m$ is the number of machines.*

**Corollary 1** *Running ROBUST-CORESET-CENTRALIZED on the output core-set of ROBUST-DISTRIBUTED produces a compact core-set of size $O\left(k + (d\log k/\delta^2)\right)$. Also, ROBUST-*

---
1. It could be any other algorithm with a constant factor approximation.

CENTRALIZED *finds a solution with* $(0.109 - \delta)$-*approximation guarantee from this compact core-set.*

The first part of Corollary 1 is a direct consequence of Theorem 1. The second part results from the approximation guarantees of Theorems 1 and 3. We refer to this version of our distributed algorithm as COMPACT-DISTRIBUTED. Next, we prove Theorem 3.

**Proof** In the first round of our algorithm, we randomly distribute the elements of $V$ on $m$ machines. i.e., independently assigning each element to one of the $m$ machines uniformly at random. The data assigned to machine $i$ is represented by $V_i$. We also define $V' = V \setminus D$ and $V'_i = V_i \setminus D$. Let $\mathcal{V}'(1/m)$ represent the distribution over random subsets of $V'$ where each element is sampled independently with a probability $1/m$.

**Lemma 5** *The distribution of* $V'_i = V_i \setminus D$ *is identical to* $\mathcal{V}'(1/m)$.

**Proof** Note that we assume the adversary does not have access to the randomness of our algorithm. Therefore, all the elements of $V \setminus D$ are distributed uniformly at random on $m$ machines. ∎

For the sake of analysis, we assume, in each run of the algorithm, for picking elements from the pool of $B_\tau$ and tie-breaking we have a fixed strict total ordering $\Pi$ of the elements of $V$. The choice of permutation $\Pi$ is uniformly at random from the symmetric group $S_n$. Indeed, we assume ROBUST-CORESET-CENTRALIZED in each round among all the elements with the marginal gain of $[\tau, (1+\epsilon)\tau)$ chooses the one with the highest rank in $\Pi$. Also, we make a slight change to the algorithm: when the size of all the elements with marginal gain in a range $[\tau, (1+\epsilon)\tau)$ is exactly $\frac{d}{\epsilon}$, we choose the element with the highest priority in $\Pi$ and pass all these elements to the next round (as part of the core-set). In this case, at most $d/\epsilon - 1$ elements can have a marginal gain in range $[\tau, (1+\epsilon)\tau)$. So, ROBUST-CORESET-CENTRALIZED would consider the next smaller threshold, i.e., elements with marginal gain in $[\frac{\tau}{1+\epsilon}, \tau)$.

Suppose $S^* = \arg\max_{S \subseteq V', |S| \leq k} f(S)$ and $f(S^*) = \text{OPT}$. In addition, let $\text{OPT}_i = \max_{S \subset V'_i, |S| \leq k} f(S)$, i.e., $\text{OPT}_i$ is the optimum value for the data on machine $i$. Let's define the set $O_i$, conditioned on the fixed set $V_i$ and the permutation $\Pi$, as follows

$$O_i = \{e \in S^* : e \notin \text{ROBUST-CORESET-CENTRALIZED}(V_i \cup \{e\})\}.$$

Note that while the output of ROBUST-CORESET-CENTRALIZED is random in general; if we assume the set $V_i$ and total ordering $\Pi$ are fixed a priori, then the set $O_i$ is deterministic also.

**Lemma 6** *Consider a fixed strict total ordering* $\Pi$ *between elements of* $V$. *For all* $e \in O'_i \subseteq O_i$ *we have*

$$e \notin \text{ROBUST-CORESET-CENTRALIZED}(V_i \cup O'_i),$$

*and*

$$\text{ROBUST-CORESET-CENTRALIZED}(V_i \cup O'_i) = \text{ROBUST-CORESET-CENTRALIZED}(V_i).$$

**Proof** In the first step, we show that the thresholds for Robust-Coreset-Centralized on sets $V_i \cup \{e\}$ and $V_i \cup O'_i$ are equal to thresholds of Robust-Coreset-Centralized on set $V_i$. First note that for $e \in O_i$, we have $f(\{e\}) \leq \Delta_d$. This is true because if $f(\{e\}) > \Delta_d$, then $e$ is picked by the algorithm as an element of the core-set (as one of the top $d+1$ singleton value elements) and it contradicts with the assumption that $e \notin \text{Robust-Coreset-Centralized}(V_i \cup \{e\})$. As $\Delta_0$ and $\Delta_d$ are the same for all sets $V_i, V_i \cup \{e\}$ and $V_i \cup O'_i$, their corresponding thresholds is the same also.

We prove the equality of the output core-sets of Robust-Coreset-Centralized on these three different sets by induction. For this reason assume, for a threshold $\tau$, the sets of elements chosen by Robust-Coreset-Centralized on both $V_i$ and $V_i \cup O'_i$ are equal so far. We show that the two instances of algorithm pick exactly the same element in the next step. Let $B_\tau$ and $B'_\tau$ denote the set of all elements with the marginal gain in the current bucket we are processing from Robust-Coreset-Centralized($V_i$) and Robust-Coreset-Centralized($V_i \cup O'_i$), respectively. We consider two main cases. If $O'_i \cap B'_\tau = \emptyset$, then the two sets $B_\tau$ and $B'_\tau$ we are processing in the runs of the algorithm are the same. If their size is strictly less than $\frac{d}{\epsilon}$, both instances output the set $B_\tau = B'_\tau$ as part of their core-set and consider the next smaller threshold. Therefore the core-sets output by the two runs of the algorithm will remain the same in this step as well, and the induction step is proved. Otherwise, there are at least $\frac{d}{\epsilon}$ elements, and the two instances choose the same element to add to $A_\tau$ because they take the element with the highest priority in $\Pi$.

Now consider the case $O'_i \cap B'_\tau = O''_i \neq \emptyset$. We consider two sub-cases in this part. Assume $|B_\tau| < \frac{d}{\epsilon}$, then for all $e \in O''_i$ there exists at most $\frac{d}{\epsilon} - 1$ elements in $V_i \setminus O_i$ with the marginal gain in the current bucket. This contradicts the fact that $e \in O_i$ because for every such $e$, the set $B_\tau \cup \{e\}$ has at most $\frac{d}{\epsilon}$ elements and therefore $e$ will be part of the core-set.

So we can focus on the sub-case $|B_\tau| \geq \frac{d}{\epsilon}$. Since for every $e \in O''_i$, element $e$ is not part of the core-set when added to $V_i$, there should be some higher priority element in $B_\tau$ than any $e \in O''_i$. This highest priority element will be picked by both runs of the algorithm. Therefore the core-sets remain the same in this step of induction as well which completes the proof. ∎

Next, we bound the marginal gain of elements of $O_i$ versus elements picked from pools of $\{B^i_\tau\}$ by Robust-Coreset-Centralized, i.e., set $A^i$.

**Lemma 7** *Consider a fixed strict total ordering $\Pi$ between elements of $V$. Let $A^i$ denote the set chosen by* Robust-Coreset-Centralized *on machine $i$. For all $e \in O_i$, we have*

1. *If $|A^i| < k$ then $\Delta_f(e|A^i) \leq \frac{\text{OPT}_i}{2k}$.*

2. *If $|A^i| = k$ then $\Delta_f(e|A^i) \leq \frac{(1+\epsilon)f(A^i)}{k} \leq \frac{(1+\epsilon)\text{OPT}_i}{k}$.*

**Proof** From Lemma 6, we know Robust-Coreset-Centralized on the sets $V_i$ and $V_i \cup O_i$ outputs the same sets.

1. From the fact that $|A^i| < k$, we conclude Robust-Coreset-Centralized has passed over all the thresholds and has not picked $e$. So we conclude $\Delta_f(e|A^i) \leq \frac{\Delta_d}{2(1+\epsilon)k} \leq \frac{\text{OPT}_i}{2k}$.

2. Denote $A^i$ by $\{e_1, \cdots, e_{|A^i|}\}$, where $e_j$ is the $j$-th element added to $A^i$. Also, define $A_j = \{e_1, \cdots, e_j\}$, i.e., $A_j$ is the first $j$ picked elements of $A^i$. We have

$$f(A^i) = \sum_{j=1}^{|A^i|} \Delta_f(e_j | A_{j-1}),$$

where $A_0 = \emptyset$. We know

$$\Delta_f(e|A^i) \stackrel{(a)}{\leq} \Delta_f(e|A_{j-1}) \stackrel{(b)}{\leq} (1+\epsilon)\Delta_f(e_j|A_{j-1})$$

The inequality $(a)$ is the direct consequence of submodularity of $f$. We prove $(b)$ by contradiction. Assume $(b)$ is not true. Then $e$ should have been taken as a part of the core-set before picking $e_j$, and this contradicts with $e$ being in $O_i$.

To sum up, we have

$$\Delta_f(e|A^i) \leq \frac{1+\epsilon}{|A^i|} \sum_{j=1}^{|A^i|} \Delta_f(e_j|A_{j-1}) \leq \frac{(1+\epsilon)f(A^i)}{k} \leq \frac{(1+\epsilon)\mathrm{OPT}_i}{k}.$$

∎

The next step is to bound $f(O_i)$ based on $f(A^i)$ and $\mathrm{OPT}_i$.

**Lemma 8** $f(O_i) \leq f(A^i) + (1+\epsilon)\mathrm{OPT}_i$.

**Proof** We have

$$f(O_i) \stackrel{(a)}{\leq} f(O_i \cup A^i) \stackrel{(b)}{\leq} f(A^i) + \sum_{e \in O_i} \Delta_f(e|A^i) \stackrel{(c)}{\leq} f(A^i) + (1+\epsilon)\mathrm{OPT}_i.$$

Inequality $(a)$ drives from the monotonicity of $f$. Inequality $(b)$ is true because $O_i \cap A^i = \emptyset$ and $f$ is submodular. Inequality $(c)$ is true from the result of Lemma 7 and the fact that $|O_i| \leq k$. ∎

Now, we can bound the expected value of $f(O_i)$ by the expect value of $f(S^i)$, where $S^i$ is the result of ROBUST-CENTRALIZED from the core-set of machine $i$. Assume set $A^{i'}$ consists of elements of $A^i$ after deletion. We have

$$\mathbb{E}_\Pi[f(O_i)] \stackrel{(a)}{\leq} \mathbb{E}_\Pi[A^i] + (1+\epsilon)\mathrm{OPT}_i$$

$$\stackrel{(b)}{\leq} \frac{(1+\epsilon)\mathbb{E}_\Pi[A^{i'}]}{1-\epsilon} + (1+\epsilon)\mathrm{OPT}_i$$

$$\stackrel{(c)}{\leq} \frac{(1+\epsilon)\mathbb{E}_\Pi[f(S^i)]}{1-\epsilon} + (1+\epsilon)\mathrm{OPT}_i \rightarrow$$

$$(\frac{1}{3} - 2\epsilon)\mathbb{E}_\Pi[f(O_i)] \stackrel{(d)}{\leq} \mathbb{E}_\Pi[f(S^i))]$$

Inequalities (a) and (b) are directly from the results of Lemma 8 and Lemma 2. We know $A^i \subseteq S^i$ and inequality (c) concludes from the monotonicity of $f$. Theorem 1 guarantees that ROBUST-CORESET-CENTRALIZED outputs a $(\frac{1-3\epsilon}{2}, d)$-robust randomized core-set. This ensures that, for every ground set $V_i$, $(\frac{1-3\epsilon}{2})\text{OPT}_i \leq \mathbb{E}_\Pi[f(S^i)]$. Inequality (d) results from this fact.

We note that the only randomness properties we need in Lemma 2 and Theorem 1 are to ensure each added element to an $A$ set has a probability of deletion of at most $\epsilon$ with linearity of expectation. With the $\Pi$ based implementation of this randomness, we achieve these properties. ∎

In the last step, we prove the approximation guarantee of ROBUST-DISTRIBUTED. Define vector $\boldsymbol{p}$ such that for $e \in V$, we have

$$p_e = \begin{cases} \mathbb{P}_{A \sim \mathcal{V}(1/m)}[e \in \text{ROBUST-CENTRALIZED}(A \cup \{e\})] & \text{if } e \in S^*, \\ 0 & \text{otherwise.} \end{cases}$$

**Lemma 9** *For $\alpha = \frac{1}{3} - 2\epsilon$ and $\beta = 1 - \frac{1}{e}$, we have*

$$\mathbb{E}[f(S^i)] \geq \alpha \mathbb{E}[f(O_i)] \geq \alpha f^-(\mathbf{1}_{S^*} - \boldsymbol{p})$$
$$\mathbb{E}[f(T)] \geq \beta \mathbb{E}[f(S^* \cap (\{\cup_i \cup_{\tau \in T^i} A_\tau^{i'}\} \cup \{\cup_i B^{i'}\}))] \geq \beta f^-((\boldsymbol{p})),$$

*where $f^-$ is the Lovász extension of function $f$.*

**Proof** The proof of this lemma is similar to the proof of (Barbosa et al., 2015, Theorem 5). Let $Z$ denote the set returned by ROBUST-DISTRIBUTED. From Lemma 9, we have

$$\mathbb{E}[f(Z)] \geq \mathbb{E}[f(S^i)] \geq \alpha f^-(\mathbf{1}_{S^*} - \boldsymbol{p}) \tag{7}$$
$$\mathbb{E}[f(Z)] \geq \mathbb{E}[f(T)] \geq \beta f^-(\boldsymbol{p}) \tag{8}$$

From the result of Eqs. (7) and (8) we have

$$(\beta + \alpha)\mathbb{E}[f(Z)] \geq \alpha\beta(f^-(\mathbf{1}_{S^*} - \boldsymbol{p}) + f^-((\boldsymbol{p}))) \overset{(a)}{\geq} \alpha\beta f^-(\mathbf{1}_{S^*}) = \alpha\beta f(S^*).$$

In inequality (a), we use the convexity of Lovász extension and (Barbosa et al., 2015, Lemma 1). This proves the first part of theorem. ∎

From Theorem 1, we know that the size of core-set for an instance of ROBUST-CORESET-CENTRALIZED is $O(k + d\frac{\log k}{\epsilon^2})$. Therefore, the size of core-set for ROBUST-DISTRIBUTED is at most $m$ times of this value.

## 5. Experimental Results

In this section, we extensively evaluate the performance of our algorithms on several publicly available real-world datasets. We consider algorithms that can be robust to the deletion of any number of items and return $k$ elements after deletion. Unfortunately, both OSU (Orlin et al., 2016) and PRO-GREEDY (Bogunovic et al., 2017) are robust to the deletion of only $o(k)$ items. For this reason, we compare our proposed methods with two other baselines:

(i) the deletion-robust algorithm designed by Mirzasoleiman et al. (2017) (we refer to it by ROBUST), and (ii) the stochastic greedy algorithm (Mirzasoleiman et al., 2015) (SG), where we first obtain a solution $S$ of size $r = 6k$ (we set $r > k$ to make the solution robust to deletion), and then we report $\text{GREEDY}(S \setminus D)$ as the final answer. In our evaluations, we consider the effect of three different parameters: (i) an algorithm is designed to be robust to deletion of $d$ elements; (ii) cardinality constraint $k$ of the final solution; and (iii) number of deleted elements $r$. The final objective value of all algorithms are normalized to the utility obtained from a classical greedy algorithm that knows the set of deleted items $D$ beforehand.

Note that we are able to guarantee the performance of our algorithms (also this is true for ROBUST (Mirzasoleiman et al., 2017)) only when the number of deletions $r$ is less than $d$. In these experiments, we are also interested in evaluating the effect of larger number of deletions, i.e., for of $r \gg d$.

### 5.1 Location Privacy

In a wide range of applications, the data can be represented as a kernel matrix $K$, which encodes the similarity between different items in the database. In order to find a representative set $S$ of cardinality $k$, a common object function is

$$f(S) = \log \det(I + \alpha K_{S,S}), \tag{9}$$

where $K_{S,S}$ is the principal sub-matrix of $K$ indexed by $S$ and $\alpha > 0$ is a regularization parameter (Herbrich et al., 2003; Seeger, 2004; Krause and Guestrin, 2005). This function is monotone submodular.

In this section, we analyze a dataset of 10,000 geolocations. Each data entry is longitude and latitude coordinates of Uber pickups in Manhattan, New York in April 2014 (UberDataset, 2016). Our goal is to find $k$ representative samples using objective function described in Eq. (9). The similarity of two location samples $i$ and $j$ is defined by a Gaussian kernel $K_{i,j} = \exp(-d_{i,j}^2/h^2)$, where the distance $d_{i,j}$ (in meters) is calculated from the coordinates and $h$ is set to 5000. We set $d = 5$, i.e., we make algorithms (theoretically) robust to deletion of at most five elements. To compare the effect of deletion of $r = |D|$ elements on the performance of algorithms,[2] we use two deletion strategies to select these items: (i) classical greedy algorithm, and (ii) the stochastic greedy algorithm.

In the first experiment, we study the effect of deleting different number of items on the normalized objective values. In order to refer to an algorithm with a specific deletion strategy, we use the name of algorithm followed by the deletion strategy, e.g., Rob-Stream-G refers to ROBUST-STREAMING where the deleted items are picked by greedy strategy. From Fig. 1a, we observe that ROBUST-STREAMING and ROBUST-CENTRALIZED are more robust to deletion than ROBUST and SG. The effect of deleting by greedy strategy on the performance of algorithms is more pronounced than SG strategy. It can be seen that, even by deleting more than $d = 5$ items, our algorithms maintain their performance. Also, SG (which is not designed to be robust to deletions) shows the worst performance. Other than normalized objective values, the memory requirement of each algorithm is quite important. Indeed, we are interested in deletion-robust algorithms that do not keep many items. Fig. 1b compares the memory complexity of algorithms. We observe that ROBUST-CENTRALIZED needs to keep the least number of items. For ROBUST algorithm, the memory complexity increases super linear in $k$ (it is $O(k \log k)$), which makes it quite impractical for large values of $k$ and $d$. To sum-up, we observe that our proposed algorithms provide the best of two worlds: while their normalized objective values are clearly better than other baselines, they need to keep much fewer number of items.

---
2. Note that in practice $r$ could be any number larger than $d$.

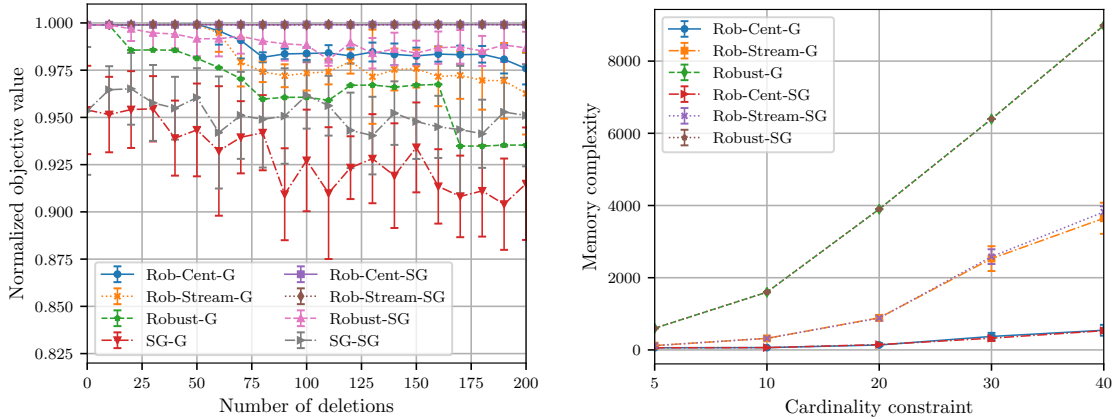

(a) We set $d = 5$ and $k = 20$.    (b) We set $d = 5$ and $r = 100$.

Figure 1: Uber dataset: (a) effect of deletion on the performance of algorithms with respect two different deletion strategies; (b) memory complexity of robust algorithms for different cardinality constraints.

## 5.2 Submodular Feature Selection

One of the challenges in learning from high dimensional data is to select a subset of relevant features in a computationally feasible way. For this reason, the quality of a subset of features $S$ can be captured by the mutual information between attributes in $S$ and the class variable $Y$ (Krause and Guestrin, 2005). More specifically,

$$I(Y; X_S) = \sum_{y \in \mathcal{Y}} \sum_{x \in \mathcal{X}_S} p(x, y) \log_2 \left( \frac{p(x, y)}{p(x)p(y)} \right),$$

where $X_S$ is a random variable that represents the set $S$ of $k$ features. The joint distribution on $(Y, X_1, \cdots, X_k)$, under the Naive Bayes assumption, is defined by $p(y, x_1, \cdots, x_k) = p(y) \prod_{i=1}^{k} p(x_i|y)$. This assumption makes the computation of joint distribution tractable. In our experiments, we estimate each $p(x_i|y)$ by counting frequencies in the dataset. In the feature selection problem, the goal is to choose $k$ features such that maximizing $f(S) = I(Y; X_S)$. It is known that the function $f(S) = I(Y; X_S)$, under the Naive Bayes assumption, is monotone submodular (Krause and Guestrin, 2005).

In this section, we use this feature selection method on two real datasets. We first show that our robust algorithms, after the deletion of sensitive features,[3] provide results with near optimal quality (based on mutual information). Second, we demonstrate that classifiers that are trained on these selected features perform very well. In our experiments, we set $d = 3$.

---

3. Features that might cause unfairness in the final classifier.

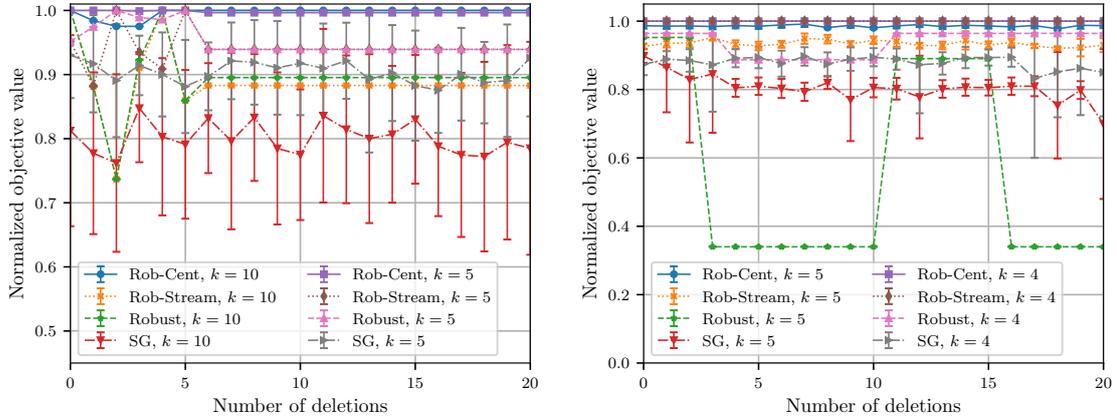

(a) Adult Income  (b) Crime and Communities

Figure 2: The effect of deletion on the performance of different algorithms for feature selection. We set $d = 3$

### 5.2.1 Fairness in Income Prediction

In this set of experiments, we use the *Adult Income* dataset from UCI Repository of machine learning databases (Blake and Merz, 1998). This dataset contains information about 32,561 individuals and whether income of those individuals is over 50K a year. We extract 113 binary features from this dataset. The goal of the classification task is to predict the income status of 16,281 test cases. For the deletions, we remove sensitive features that might result in the unfairness, e.g., features about sex, race, nationality, marital status and relationship status.

Fig. 2a compares algorithms based on different number of deletions for $k = 5$ and $k = 10$. We observe that for both values of $k$, Robust-Centralized considerably outperforms Robust (Mirzasoleiman et al., 2017) and SG. Also, the performance of Robust is better than SG.

To further investigate the effect of deletions, we compare accuracy of different classifiers, where each is trained on the features found by our algorithms and other baselines. We train two type of classifiers: (i) Naive Bayes (Zhang, 2004) and (ii) SVM (Smola and Schölkopf, 2004). From Table 2, we observe that a SVM classifier, which is trained over all features, results in an accuracy of 83.0%. If we use a greedy algorithm to find the best 5 features and train SVM classifier on those features, the accuracy will drop to 79.6%.[4] After deleting 10 features that might result in unfairness in classification (e.g., race and sex), we again use the greedy algorithm to find the best five features (referred to as GREEDY$_D$). The accuracy in this case is 79.3%. Also, interestingly, we observe that the accuracies of classifiers which are trained on the features found by Robust-Centralized and Robust-Streaming drop only by 0.2%. Furthermore, for the Naive Bayes classifier, we do not observe any decrease on the accuracy when we train on the features found by our algorithms. Finally, both Centralized (22) and Streaming (29) algorithms need to keep fewer number of items than ROBUST (39).

---
4. Clearly there is a trade off between the number of features and accuracy.

Table 2: The comparison of different classifiers (Naive Bayes and SVM for Adult Income dataset and RIDGE for Crime and Communities dataset). Ten sensitive features are deleted. The number of stored features is reported in parenthesis.

| Algorithm | Naive Bayes (Acc.) | SVM (Acc.) | RIDGE (RMSE) |
|---|---:|---:|---:|
| All features | 0.798 | 0.830 | 0.136 |
| GREEDY | 0.788 | 0.796 | 0.193 |
| GREEDY$_D$ | 0.781 | 0.793 | 0.199 |
| Rob-Cent | 0.781 (22) | 0.791 | 0.163 (25) |
| Rob-Stream | 0.781 (29) | 0.791 | 0.177 (52) |
| ROBUST | 0.779 (39) | 0.788 | 0.197 (58) |

5.2.2 FAIRNESS IN CRIME RATE PREDICTION

In the second experiment for robust feature selection, we use the *Communities and Crime* dataset from UCI Repository of machine learning databases (Blake and Merz, 1998). This dataset consist of 122 features with plausible connection to crime in communities within the United States. The crime rate is provided as the per capita violent crimes. In this experiment, we delete sensitive features such as distribution of race and sex in population and police forces. Fig. 2b compares normalized objective values for $k \in \{4, 5\}$ and different number of deletions. Again, we observe that our centralized and streaming algorithms have the best performances. We should point out that the parameter $d$ can also play an important role in practice. Indeed, since all algorithms are made robust to deletion of $d = 3$ elements, the performance of ROBUST (Mirzasoleiman et al., 2017) hugely decreases with only $r = 4$ deletions, while our algorithms maintain their near optimal performances.

To assess the quality of selected features, we use a RIDGE regression classifier (Hoerl and Kennard, 1970). The RMSE for a classifier that is trained on all features is 0.136. For classifiers trained on features selected by GREEDY and GREEDY$_D$, the errors increase to 0.193 and 0.199, respectively. The errors for centralized (0.163) and streaming (0.177) algorithms are even less than the greedy algorithm which knows the deleted features in advance. This might be due to the fact that only our proposed methods select features related to the percentage of divorced males and females as important attributes. It is plausible that these attributes can have high correlations with crime rate.

**5.3 Distributed Algorithm and Massive Data Summarization**

To evaluate the performance of ROBUST-DISTRIBUTED on massive datasets, we consider the *Census1990* dataset from UCI Repository of machine learning databases (Blake and Merz, 1998). This dataset is consists of 2,458,285 data points with 68 features. We are going to find $k$ representative samples from this large dataset. We apply the set selection objective function described in Eq. (9). The similarity between two entries $x$ and $x'$ is defined by $1 - \frac{\|x-x'\|}{\sqrt{68}}$, where $\|x - x'\|$ is the Euclidean distance between feature vectors of $x$ and $x'$.

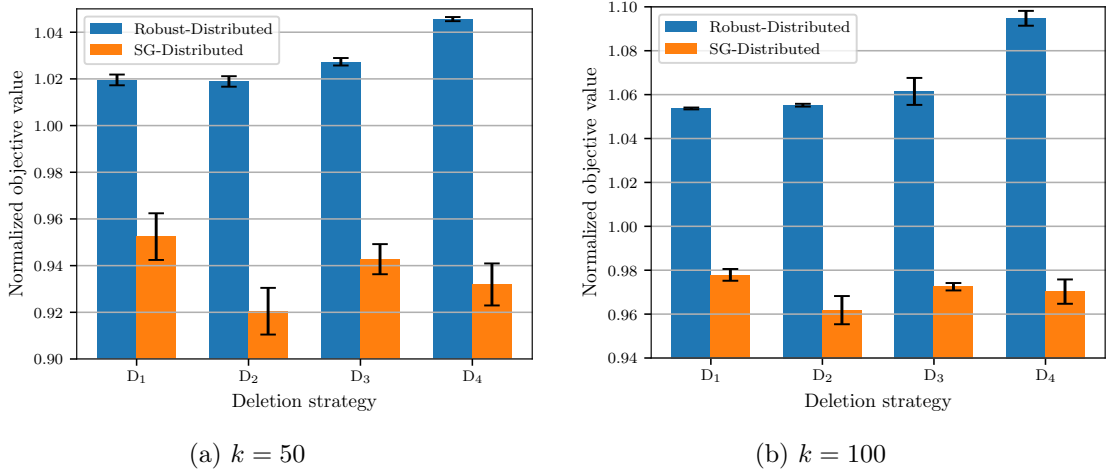

(a) $k = 50$

(b) $k = 100$

Figure 3: Census1990 dataset: Robust-Distributed versus SG-Distributed for four different deleting strategies.

We randomly split the dataset into $m = 12$ partitions. For each instance of Robust-Coreset-Centralized, we set $d = 25$ with an $\epsilon = 0.1$. As a baseline, we consider a distributed version of stochastic greedy algorithm (refer to it as SG-Distributed). For this algorithm, we first run stochastic greedy on each partitions to select $S_i = 6k$ items. After deletion of $D$, we report $f(\text{GREEDY}(\cup S_i \setminus D))$ as the final result. Also, we normalize the utility of functions to the objective value of an instance of SG-Distributed that knows the set of deleted items $D$ in advance. For deletions, we propose four different strategies: $D_1$ randomly deletes 50% of items, $D_2$ randomly deletes 80% of items, $D_3$ deletes all men in the dataset, and $D_4$ deletes all women in the dataset.

We investigate the effect of different deletion strategies for two values of $k \in \{50, 100\}$. In Figs. 3a and 3b, we observe that Robust-Distributed clearly outperforms SG-Distributed in all cases. Furthermore, we observe that the objective value of Robust-Distributed in all scenarios is even better than our reference function for normalization (normalized objective values are larger than 1). Each machine on average stores 209.3 (for $k = 50$) and 348.3 (for $k = 100$) items. The standard deviations of memory complexities are 36.9 and 26.5, respectively. To conclude, Robust-Distributed enables us to robustly summarize a dataset of size 2,458,285 with storing only ≈4500 items. Our experimental results confirm that this core-set is robust to the deletion of even 80% of items.

## 6. Conclusion

In this paper, we considered the problem of deletion-robust submodular maximization. We provided the first scalable and memory-efficient solutions in different optimization settings, namely, centralized, streaming, and distributed models of computation. We rigorously proved that our methods enjoy constant factor approximations with respect to the optimum algorithm that is also aware of the deleted set of elements. We showcased the effectiveness of our algorithms on real-word problems where part of data should be deleted due to privacy and fairness constraints.